\documentclass[11pt,a4paper]{article}
\usepackage[hyperref]{acl2021}
\usepackage{times}
\usepackage{latexsym}
\usepackage{graphicx}
\usepackage{tabularx}

\usepackage{microtype}

\aclfinalcopy % Uncomment this line for the final submission
\newcounter{notecounter}

\newcommand{\enoteson}{\long\gdef\enote##1##2{{
            \stepcounter{notecounter}
            \large\bf
            \hspace{100cm}\arabic{notecounter} $<<<$ ##1: ##2
            $>>>$\hspace{1cm}}}}
\enoteson
\title{Anchor-based Bilingual Word Embeddings for Low-Resource Languages}

\author{Tobias Eder$^1$, Viktor Hangya$^2$ \and Alexander Fraser$^2$ \\
  $^1$Research Group Social Computing \\ Technical University of Munich \\
  $^2$Center for Information and Language Processing \\
  LMU Munich \\
  \texttt{tobias.eder@in.tum.de}, \\
  {\tt \{hangyav, fraser\}@cis.lmu.de}
}

\date{}

\begin{document}
\maketitle
\begin{abstract}
Good quality monolingual word embeddings (MWEs) can be built for languages which have large amounts of unlabeled text. MWEs can be aligned to bilingual spaces using only a few thousand word translation pairs.
For low resource languages training MWEs monolingually results in MWEs of poor quality, and thus poor bilingual word embeddings (BWEs) as well.
This paper proposes a new approach for building BWEs in which the vector space of the high resource source language is used as a starting point for training an embedding space for the low resource target language. By using the source vectors as anchors the vector spaces are automatically aligned during training. 
We experiment on English-German, English-Hiligaynon and English-Macedonian.
We show that our approach results not only in improved BWEs and bilingual lexicon induction performance, but also in improved target language MWE quality as measured using monolingual word similarity.
\end{abstract}

\section{Introduction}
\label{into}

Bilingual Word Embeddings are useful for cross-lingual tasks such as cross-lingual transfer learning or machine translation.
Mapping based BWE approaches rely only on a cheap bilingual signal, in the form of a seed lexicon, and monolingual data to train monolingual word embeddings (MWEs) for each language, which makes them easily applicable in low-resource scenarios \cite{Mikolov2013exploiting,Xing2015,Artetxe2016}.
It was shown that BWEs can be built using a small seed lexicon \cite{Artexte2017} or without any word pairs \cite{Conneau2018,Artetxe2018} relying on the assumption of isomorphic MWE spaces.
Recent approaches showed that BWEs can be built
%jointly on the source and target language monolingual corpora
without the mapping step.
\citet{Lample2018SMT} built \textsc{FastText} embeddings \cite{bojanowski-etal-2017-enriching} on the concatenated source and target language corpora exploiting the
%joint
shared
character n-grams in them.
Similarly, the shared source and target language
% Byte-Pair Encoding (BPE)
subword tokens are used as a cheap cross-lingual signal in \citet{devlin-etal-2019-bert,Conneau2019XLM}.
Furthermore, the advantages of mapping and
%joint approaches were
jointly training the MWEs and BWEs were
%combine
combined
in \citet{Wang2020} for even better BWEs.

While these approaches already try to minimize the amount of bilingual signal needed for cross-lingual applications, they still require a larger amount of monolingual data to train semantically rich word embeddings \cite{adams-etal-2017-cross}.
This becomes a problem when one of the two languages does not have sufficient monolingual data available \cite{artetxe2020rigor}.
In this case, training a good embedding space can be infeasible which means mapping based approaches are not able to build useful BWEs \cite{Michel2020}.

In this paper we introduce a new approach to building BWEs when
one of the languages only has limited available monolingual data.
Instead of using mapping or joint approaches, this paper takes the middle ground by making use of the MWEs of a resource rich language and training the low resource language embeddings on top of it. For this, a bilingual seed lexicon is used to initialize the representation of target language words by taking the pre-trained vectors of their source pairs prior to target side training, which acts as an informed starting point to shape the vector space during the process.
We randomly initialize the representations of all non-lexicon target words and
run Continuous Bag-of-Words (CBOW) and skip-gram (SG) training procedures to
generate target embeddings with both \textsc{Word2Vec}
\cite{Mikolov2013efficient} and \textsc{FastText}
\cite{bojanowski-etal-2017-enriching}.
Our approach ensures that the source language MWE space is intact,
so that the data deficit on the target side does not result in lowered source embedding quality.
The improved monolingual word embeddings for the target language
outperform embeddings trained solely on monolingual data
for semantic tasks such as word-similarity prediction.
We study low-resource settings for English-German and English-Hiligaynon, where previous approaches have failed \cite{Michel2020}, as well as English-Macedonian.

\section{Method}
\label{sec:method}

Previous mapping approaches rely on the alignment of two pre-trained monolingual word embedding spaces.
%% AF: could put this back later
%This naturally limits the resulting mapping by the monolingual performance, or expressiveness, of either space involved in the mapping.
In case one of the two languages has significantly fewer resources available, this will strongly affect the resulting mapping negatively.
This is also an issue for joint approaches because
%in case the source and target corpora sizes are imbalanced,
the shared token representations are biased towards the language with more training samples.
%Instead
%the
%our
%idea is to leverage
%the additional resources available for
Our approach instead leverages
the high resource language to improve
%overall
performance on the low-resource language.

%Our approach functions in the following way: Instead of training two separate MWEs on monolingual data or training BWEs jointly on them,
We pre-train MWEs for the source language and use the source MWEs to initialize the space of the low resource target language.
Using a set of initial seed pairs, the representation of a seed word in the target space is replaced with the representation of its translation (anchor points).
Then, training is performed on the initialized space using only monolingual data from the low resource language by only updating the representation of non-seed words which are initialized randomly.
Through this method a BWE representation is directly induced from the anchor points of the fixed vectors.

In some cases there are multiple valid translations for a single target language word. We experiment with either initializing with the average over these possible translations or randomly selecting only one of them. The averaging helps by finding a common anchor for the different semantic nuances the token might represent in different
%context of the target language.
target language contexts.
Additionally, we experiment with enabling or disabling the updates of anchor vectors during training.
We implemented the anchor point based initialization in both \textsc{Word2Vec} and \textsc{FastText} with only complete token representations serving as potential anchors.
In the case of \textsc{FastText} these initializations have no influence on the subword (character n-gram) embeddings which are still initialized randomly, which makes intuitive sense in the common case of morphologically different language pairs.
Training is performed using standard hyperparameters included in the \textsc{gensim Word2Vec} and \textsc{FastText} packages \cite{rehurek_lrec}.
Unless stated otherwise, vectors are of dimensionality 300 with a context window of 5 words used during training. All models are trained for 5 epochs without further hyperparameter tuning utilizing a single desktop machine on a Intel Core i7-7700K CPU with 4.20Ghz, a NVIDIA GeForce GTX 1080 Ti graphics card and 32 GB of DDR4 SDRAM. The parameters of each trained model are equal to the standard implementation of the packages as listed above. Training time is largely dependent on input size, but corresponds to a few seconds up to roughly 5 minutes in the low resource setting.

\subsection{Experimental Setup}
\label{sec:setup}

First we conduct experiments on the German and English language pair,
since large available corpora made it easier to test different sized dictionaries and corpora during training. The basic setup trains a MWE on the source language (English) up front. For this training the WMT 2019 News Crawl corpus in English, including approximately 532 million tokens, was chosen \cite{Barrault2019}.
Similarly for the target language, we used the German WMT 2019 News Crawl from which we uniformly sample to obtain training sets of different sizes.
All dataset are tokenized and lowercased before training.

%We
To evaluate, we
translate German words to English. We use
%As a seed lexicon for training and evaluation,
the MUSE German-English dictionary
%was chosen
\cite{Conneau2018}.
%The complete dictionary includes
There are
102K
%thousand
translation pairs with a total of roughly
68K
%thousand
unique German words.
For each German word there might be multiple valid English translations, which are listed in the dictionary.
For the initialization we select either randomly one translation option or the averaged word representations of all available translations, as discussed in section~\ref{sec:method}.
However,
%a large number of vocabulary only has
many German words have only
one valid translation.
%As the test set w
We used the MUSE test set containing roughly 3000 translation pairs in the frequency range 5000-6500, leaving 99K
%thousand
%translation
pairs as potential candidates for the initialization.
In our experiments we mostly consider setups with much smaller training lexicon sizes, by taking the top-$n$ most frequent source words and their translations from the lexicon.

In addition to the German experiments we test our system on two lower resource languages: Macedonian and Hiligaynon.
For Macedonian we use data in the form of a Wikipedia dump, as well as the MUSE dictionary for the language pair Macedonian-English for our test setup.\footnote{https://dumps.wikimedia.org/mkwiki/ (downloaded on 01/31/21)}
For Hiligaynon we use a corpus containing roughly 350K tokens as well as a corresponding dictionary containing 1100 translated terms between English and Hiligaynon and an additional test set of 200 terms released by \citet{Michel2020}.
%% AF: removed, this is mentioned below
%, comparing directly with their results.
%We use the same resources to compare our method to the one presented in that paper.

After training, bilingual lexicon induction (BLI) is done by taking the top $n$
closest vectors measured by cross-domain similarity local scaling (CSLS)
% NN
distance.
%% AF could be put back later (but rewrite)
%from the representation of a word in one language to the closest corresponding vectors in its position in the original MWE.
For better comparability we use the evaluation method provided by MUSE \citep{Conneau2018} for both the comparison baseline as well as our system. For Hiligaynon we use cosine to compare directly with \cite{Michel2020}.
%In the Hiligaynon case the evaluation setup mirrors the one presented in \cite{Michel2020} using cosine distance for direct comparability to that paper.

\section{Results}
\label{sec:results}

The following section evaluates different models quantitatively using $acc@5$
and $acc@1$ as a metric. The baseline runs MUSE tool in supervised mode using
iterative procrustes refinement
% \citep{Conneau2018}
to obtain the mapping
% The baseline runs
using default parameters
% for MUSE
as reported in \citet{Conneau2018}. For the English embedding the full corpus size was used, while in the case of the (low-resource) languages the corpora sizes were varied to observe changes in performance.

    \begin{table*}
      \centering
      \scalebox{.7}{
        \begin{tabular}{lrrrrrrrrr}
        \hline
        \multicolumn{1}{c}{\textbf{Model // Corpus Size}} &
        \multicolumn{1}{c}{\textbf{100K}} &
        \multicolumn{1}{c}{\textbf{300K}} &
        \multicolumn{1}{c}{\textbf{500K}} & 
        \multicolumn{1}{c}{\textbf{1M}} &  \multicolumn{1}{c}{\textbf{2M}} & 
        \multicolumn{1}{c}{\textbf{5M}} & 
        \multicolumn{1}{c}{\textbf{10M}} & 
        \multicolumn{1}{c}{\textbf{20M}} & 
        \multicolumn{1}{c}{\textbf{50M}} \\
        \hline \hline
        \multicolumn{1}{l}{Baseline Word2Vec (CBOW)} &0.0 (0.0)&0.0 (0.0)&0.0 (0.0)& 0.3 (0.0) & 0.6 (0.2) & 6.1 (1.4) & 15.9 (4.2) & 26.7 (12.8) & 40.2 (21.3) \\
         \multicolumn{1}{l}{Baseline FastText (SG)} &0.0 (0.0)&0.0 (0.0)&0.0 (0.0)& 0.6 (0.2) & 1.9 (0.5) & 8.9 (3.9) & 19.7 (8.6) & 32.1 (18.1) & 45.1 (28.9) \\
        \hline
        \multicolumn{1}{l}{Fixed not Averaged (CBOW)} & 1.3 (0.9) & 0.8 (0.3) & 0.8 (0.3) & 1.5 (0.5) & 4.1 (1.7) & 13.6 (5.6) & 23.9 (12.2) & 26.3 (13.4) & 35.7 (21.0) \\
        \multicolumn{1}{l}{Fixed and Averaged (CBOW)} & 1.3 (0.4) & 1.2 (0.1) & 1.8 (0.2) & 1.8 (0.6) & 4.3 (1.9) & 13.1 (5.5) & 23.3 (11.9) & 33.2 (18.4) & 44.1 (26.3) \\
        \multicolumn{1}{l}{Trained not Averaged (CBOW)} & 1.3 (0.9) & 0.8 (0.4) & 1.4 (0.8) & 3.3 (0.5) & 7.3 (1.7) & 18.0 (5.6) & 27.0 (12.2) & 35.8 (21.0) & 45.2 (28.4) \\
        \multicolumn{1}{l}{Trained and Averaged (CBOW)} & 1.7 (0.4) & 1.5 (0.5) & 2.9 (1.4) & 4.5 (1.3) & 10.7 (4.8) & 22.3 (11.9) & 32.2 (18.3) & 40.5 (25.0) & 48.5 (31.5) \\
        \end{tabular}
      }
\caption{Evaluation of models at varying data-sizes for English-German. Baselines (MUSE) and proposed methods, reporting acc@5 (acc@1).}
      \label{tab:baseline}
    \end{table*}

\subsection{Bilingual experiments}
\label{sec:results:bilin}

BLI was performed using the method from section~\ref{sec:method}. Since Word2Vec
SG and FastText embeddings performed much worse with the anchored training,
all following numbers report Word2Vec CBOW embeddings.

Table~\ref{tab:baseline} shows the comparison between four different possible setups for the proposed method as explained in Section~\ref{sec:method}: Either fixing anchor-vectors or allowing them to train or initializing with single word vectors or averaged ones. The overall best performing model utilizes averaged and non-fixed anchor vectors.
Table~\ref{tab:baseline} also shows the baselines at varying corpora sizes. Overall the anchor method performs much better than the baseline at lower corpora sizes and stays competitive as corpus size increases.
Results are similarly consistent when looking at either $acc@5$ or $acc@1$.

\begin{figure}[t]
        \includegraphics[scale=.45]{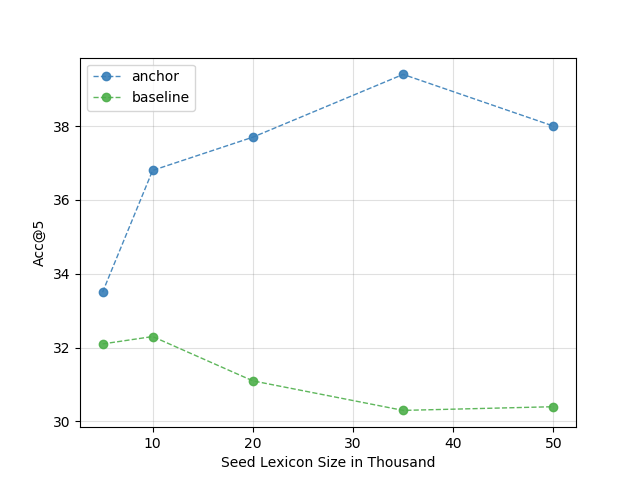}
        \caption{Anchor method for English-German with fixed vectors and baseline with varying training-dictionary sizes at corpus size 20 million
}
    \label{fig:5}
\end{figure}

One important parameter for the proposed method is how it scales with the available amount of anchor-vectors used for training. In a range of experiments, different initialization sizes were compared. Figure~\ref{fig:5} shows the result for varying the number of anchor vectors. The general trend is that the more anchor vectors, the better the performance, which slowly caps off at the higher end, as more vectors of lower frequency words start introducing noise.
The same development is not true for the baseline, which does not benefit equally from increasing the potential seed lexicon vocabulary and even starts losing performance at larger dictionary sizes. 

This difference is likely rooted in the inclusion of less frequent word pairs in
the larger dictionaries.
These words have worse quality representations which introduces noise in the
mapping process, thus restricting the precision of their orthogonal alignment as
described by \cite{Sogaard2018}.
In contrast, our method initializes all target language word embeddings given
their pairs, i.e., perfectly aligning all words in the training dictionary,
which serve as high quality anchor points for the remaining words.

%Do we need to state this at this point? I think it can go.
%Despite the anchor method performing generally better with more vectors, we kept the anchor lexicon fixed at a size of 5k, since it might be a more unrealistic setting to expect a low amount of monolingual training data in conjunction with a large bilingual dictionary. However these experiments have confirmed the positive impact of a medium amount of quality translation pairs, as evident in the steeper incline between the 5000 and 10000 seed vector mark.

\subsection{Macedonian and Hiligaynon}
\label{sec:results:oth}

Another set of experiments was done on the language pair English-Macedonian, a language that already offers less resources than German and is also more dissimilar from English.

    \begin{table*}
      \centering
      \scalebox{.7}{
        \begin{tabular}{lrrrrrr}
        \hline
        \multicolumn{1}{c}{\textbf{Model // Corpus Size}} &
        \multicolumn{1}{c}{\textbf{1M}} &  \multicolumn{1}{c}{\textbf{2M}} & 
        \multicolumn{1}{c}{\textbf{5M}} & 
        \multicolumn{1}{c}{\textbf{10M}} & 
        \multicolumn{1}{c}{\textbf{20M}} & 
        \multicolumn{1}{c}{\textbf{37M}} \\ \hline \hline
        \multicolumn{1}{l}{Baseline Word2Vec (CBOW)} & 0.0 (0.0) & 0.7 (0.0) & 6.0 (1.9) & 18.1 (7.9) & 28.0 (15.1) & 37.0 (20.4) \\
        \hline
        \multicolumn{1}{l}{Trained and Averaged (CBOW)} & 1.6 (0.4) & 4.7 (1.6) & 13.1 (5.6) & 26.3 (13.4) & 35.5 (18.4) & 44.7 (24.2) \\
        \end{tabular}
    }
\caption{Anchor method vs. baseline (MUSE) at varying data sizes reporting acc@5 (acc@1) for English-Macedonian.}
      \label{tab:mace}
    \end{table*}

Results for experiments comparing between the MUSE baseline and the anchor method are shown in Table~\ref{tab:mace}. The best performing model again combines averaged initialization with trainable anchors.

Compared to the previous experiments with German, results for Macedonian are similar, while the baseline model is overall weaker than before, suggesting the anchor method benefits more strongly from a high-resource embedding, even when language pairs become more dissimilar.

% We can cut this shorter by only looking at CBOW as well and getting rid of the SG results, since we dont report them anywhere else. Thoughts?
For English-Hiligaynon previous approaches failed due to limitations of the available monolingual training data. Table~\ref{tab:hil} shows the performance for translating Hiligaynon words to English. The evaluation was done using cosine-distance for better comparability between the \citet{Michel2020} paper and our results.
Since there were only single translations of words in the provided dictionary, the method of averaging vectors for initialization was not used. Similarly, during evaluation only one valid term per word was possible.
While \citet{Michel2020} reported $0.5\%$ for 50 dimensional vectors,
%, the same could not be repeated in our case, with
in our baseline
the 50 dimensional vectors
%achieving
achieved
a constant 0 (not shown).
The numbers are comparable to the low frequency experiments between German and English as seen in Table~\ref{tab:baseline}.
% When we cut out SG this part can also go.
%The fact that multiple models achieved the same score in these experiments was due to the low number of words being evaluated. In fact the CBOW and SG models made different errors and got different vocabulary words correct, but by coincidence achieved the same number of correctly translated terms.

\begin{table}[t]
  \centering
  \scalebox{.7}{
    \begin{tabular}{lc}
    \hline
    \multicolumn{1}{c}{\textbf{Algorithm}} &  \multicolumn{1}{c}{\textbf{Acc@5}} \\ \hline \hline
    \multicolumn{1}{l}{Baseline Michel et al. 50D} & 0.5 \\
    \multicolumn{1}{l}{Baseline Michel et al. 300D} & 0.0 \\ \hline
    \multicolumn{1}{l}{Anchor fixed not averaged CBOW 300D} & 2.6 \\
    %\multicolumn{1}{l}{Anchor fixed not averaged SG 300D} & 1.7 \\ \hline
    \multicolumn{1}{l}{Anchor trained not averaged CBOW 300D} & 4.3 \\
    %\multicolumn{1}{l}{Anchor trained not averaged SG 300D} & 4.3 \\
    \end{tabular}}
  \caption{BLI on Hiligaynon-English.}
  \label{tab:hil}
\end{table}

\subsection{Monolingual experiments}
\label{sec:results:mono}

\begin{figure}[t]
        \includegraphics[scale=.45]{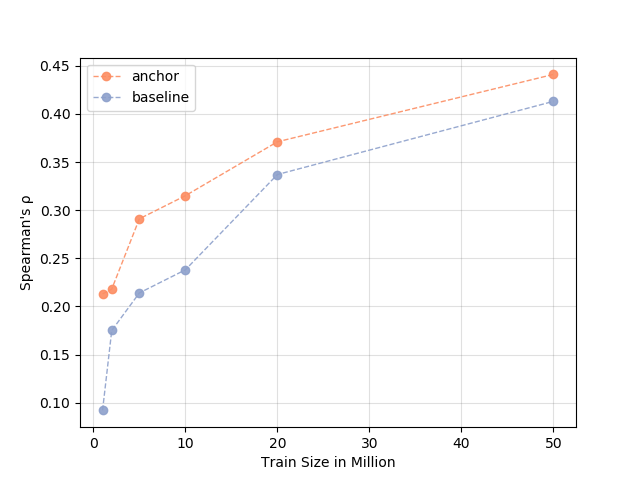}
        \caption{Spearman's rho correlation on monolingual word similarity across corpora sizes for German.}
    \label{fig:7}
\end{figure}

In addition to better BWEs, our approach also improves the low-resource embedding for purely monolingual tasks.
To confirm this, the anchor-vector trained embeddings for German were evaluated on monolingual word similarity and compared to the results achieved by regular training of the embedding space.
We evaluate on multiple datasets:
GUR350 and GUR65 \citep{gurevych2005using}, SEMEVAL17 \citep{cer2017semeval}, SIMLEX-999 \citep{leviant2015separated}, WS-353 \citep{agirre2009study} and ZG22 \citep{zesch2006automatically},
and report the averaged Spearman's rho correlation between cosine similarity of vector pairs and human annotations. 
%Should we mention this? One reviewer was complaining about us only evaluating monolingual on German in the long paper
%Unfortunately s
Similar monolingual
%tasks
datasets
are 
%currently
not available for
%either
Macedonian and Hiligaynon.
In Figure~\ref{fig:7} the effect of employing the anchor method on monolingual word similarity performance is compared against Word2Vec CBOW trained without anchor initialization.
The improvements across different training corpora sizes are in favor of the proposed method, suggesting that it can be employed to improve performance on monolingual tasks.
Overall this serves to demonstrate the advantage of the anchor-method on small datasets and allows to learn better monolingual representation from the same amount of data by utilizing the information from a pretrained embedding for a completely different language with more readily available training data. The thus learned representation can not only serve as an already aligned space for translation tasks as shown above, but is also the better performing representation of the monolingual space.
%, without relying on extra data from the language.

\section{Conclusion}
\label{sec:con}

We proposed a novel approach to build BWEs to improve performance on language pairs with limited monolingual data on the target side.
By utilizing pre-trained MWEs of resource rich languages and
a seed lexicon to fix anchor points before training, a structurally similar embedding space can be learned for the low resource language which is aligned with the source representations.
We evaluated our approach on the BDI task using English-German to test varying training parameters and corpora sizes, on English-Macedonian and the extremely low resource language pair English-Hiligaynon on which previous approaches failed.
We showed that the performance of existing mapping approaches degrades drastically with lower monolingual data sizes, even when there are large seed lexicons available.
In contrast, our proposed system outperformed previous mapping based approaches on these setups including English-Hiligaynon.
On top of improved BWEs, we showed improved MWE quality as well for the target language by outperforming standard MWEs on the monolingual word similarity task showing that it is beneficial for monolingual tasks as well.
We implemented our approach for both Word2Vec and FastText which we publicly
release to promote reproducibility and further
research.\footnote{http://cistern.cis.lmu.de/anchor-embeddings}

\section*{Ethical Considerations}

The proposed system acts as a tool to specifically help in the low resource setting that predominantly affects less researched languages. Even though part of the experiments were done on the higher resource language pair English-German, the results were further confirmed for other pairs of languages.

As a word embedding based system, the resulting mappings and embedding spaces are highly affected by the kind of monolingual content that goes into their training, which is why we made sure to train the embeddings on texts that should adhere to a higher standard, such as verified news media and online articles, instead of a general web crawl. Additionally the seed lexicons used come from verified sources, such as the popular MUSE lexicons in the case of English-German and English-Macedonian as well as from translations by a native speaker of Hiligaynon in the case of English-Hiligaynon.

We hope that in general the proposed methods can help alleviating some of the resource problems less researched languages are facing and thus to close the gap for language technology working with and on these languages.

As part of the ethical responsibility to ensure reproducibility and responsibility in terms of computational resources, we reported results with
a set of standard hyperparameters instead of searching for the most optimal setting for our proposed method. Our models are as lightweight as regular training methods for word embeddings and are therefore not very demanding in terms of computation. This is especially true in the low-resource setting, where training time is reduced to just a fraction compared to the bigger corpora.

\section*{Acknowledgments}

The work was supported by the European Research Council (ERC) under the
European Union’s Horizon 2020 research and innovation programme (grant
agreement No.~640550) and by German Research Foundation (DFG; grant FR
2829/4-1).

\bibliographystyle{acl_natbib}
\bibliography{anthology,acl2021}

%\appendix
%\newpage
%\clearpage
% This will be submitted in a separate pdf and cut from this version

\end{document}